\documentclass[10pt,twocolumn,letterpaper]{article}

\usepackage{cvpr}      % To produce the REVIEW version

\usepackage{pifont}
\usepackage{enumitem}
\usepackage[dvipsnames]{xcolor}
\definecolor{cvprblue}{rgb}{0.21,0.49,0.74}
\usepackage[pagebackref,breaklinks,colorlinks,citecolor=cvprblue]{hyperref}
\usepackage{url}                    % simple URL typesetting
\usepackage{booktabs}               % professional-quality tables
\usepackage{amsfonts}               % blackboard math symbols
\usepackage{amsmath}
\usepackage{color,colortbl}
\usepackage{nicefrac}               % compact symbols for 1/2, etc.
\usepackage{microtype}              % microtypography
\usepackage{multirow}
\usepackage{pgfplots}
\usepackage[accsupp]{axessibility} % Improves PDF readability for those with visual impairments.
\DeclareMathOperator*{\argmax}{argmax}
\DeclareMathOperator*{\argmin}{argmin}
\newcommand{\methodname}{$T3AL$}

\newcommand{\numframes}{N}
\newcommand{\numpredictions}{M}
\newcommand{\lossx}{Representation loss}
\newcommand{\losss}{Separation loss}
\DeclareMathOperator{\diag}{diag}
\newcommand{\video}{\mathcal{V}}
\newcommand{\image}{x}

\newcommand{\actionclassset}{\mathcal{C}}

\newcommand{\trainingset}{\mathcal{D}_{train}}

\newcommand{\testset}{\mathcal{D}_{test}}
\newcommand{\model}{\mathcal{M}}
\newcommand{\visionencoder}{\mathcal{E}_V}
\newcommand{\textencoder}{\mathcal{E}_L}
\newcommand{\positiveset}{\mathcal{Z}^+}
\newcommand{\encoderprojector}{\mathcal{P}}
\newcommand{\negativeset}{\mathcal{Z}^-}

\newcommand{\score}{s}

\def\ie{\textit{i.e}\onedot}

\newcommand{\inlineColorbox}[2]{\begingroup\setlength{\fboxsep}{1pt}\colorbox{#1}{\hspace*{2pt}\vphantom{Ay}#2\hspace*{2pt}}\endgroup}
\definecolor{VisionBlue}{RGB}{218,232,252}
\definecolor{LanguageOrange}{RGB}{255,230,204}
\definecolor{ModelGreen}{RGB}{213,232,212}
\definecolor{FunctionPurple}{RGB}{225,213,231}
\definecolor{ParamRed}{RGB}{248,206,204}
\definecolor{MoreVividModelGreen}{RGB}{173,223,172}
\definecolor{MoreVividFunctionPurple}{RGB}{192,134,216}
\definecolor{MoreVividLanguageOrange}{RGB}{255,184,115}
\definecolor{LineBlue}{RGB}{045, 114, 255}
\definecolor{MoreVividVisionBlue}{RGB}{165,232,255}

\pgfplotsset{compat=1.18} 

\def\paperID{10264} % *** Enter the Paper ID here
\def\confName{CVPR}
\def\confYear{2024}

\title{Test-Time Zero-Shot Temporal Action Localization}
\author{
{Benedetta Liberatori\textsuperscript{1}
\quad
Alessandro Conti\textsuperscript{1}
\quad
Paolo Rota\textsuperscript{1}
\quad
Yiming Wang\textsuperscript{2}
\quad
Elisa Ricci\textsuperscript{1,2}}\\[-4mm]\and
{University of Trento\textsuperscript{1}\quad Fondazione Bruno Kessler\textsuperscript{2}}\\[1mm]
{\tt\small\url{https://github.com/benedettaliberatori/T3AL}}
}

\begin{document}
\maketitle
\begin{abstract}
Zero-Shot Temporal Action Localization (ZS-TAL) seeks to identify and locate actions in untrimmed videos unseen during training. Existing ZS-TAL methods involve fine-tuning a model on a large amount of annotated training data. While effective, training-based ZS-TAL approaches assume the availability of labeled data for supervised learning, which can be impractical in some applications. Furthermore, the training process naturally induces a domain bias into the learned model, which may adversely affect the model's generalization ability to arbitrary videos. These considerations prompt us to approach the ZS-TAL problem from a radically novel perspective, relaxing the requirement for training data. To this aim, we introduce a novel method that performs \textbf{T}est-\textbf{T}ime adaptation for \textbf{T}emporal \textbf{A}ction \textbf{L}ocalization (\textbf{\methodname}). In a nutshell, \methodname~adapts a pre-trained Vision and Language Model (VLM).
\methodname~operates in three steps. First, a video-level pseudo-label of the action category is computed by aggregating information from the entire video. Then, action localization is performed adopting a novel procedure inspired by self-supervised learning. Finally, frame-level textual descriptions extracted with a state-of-the-art captioning model are employed for refining the action region proposals.  
We validate the effectiveness of \methodname~by conducting experiments on the THUMOS14 and the ActivityNet-v1.3 datasets. Our results demonstrate that \methodname~significantly outperforms zero-shot baselines based on state-of-the-art VLMs, confirming the benefit of a test-time adaptation approach.
\end{abstract}
  
\vspace{-10pt}
\section{Introduction}
\label{sec:intro}

\begin{figure}[!t]
    \centering
    \subfloat[Previous approaches]{{\includegraphics[width=0.9\linewidth]{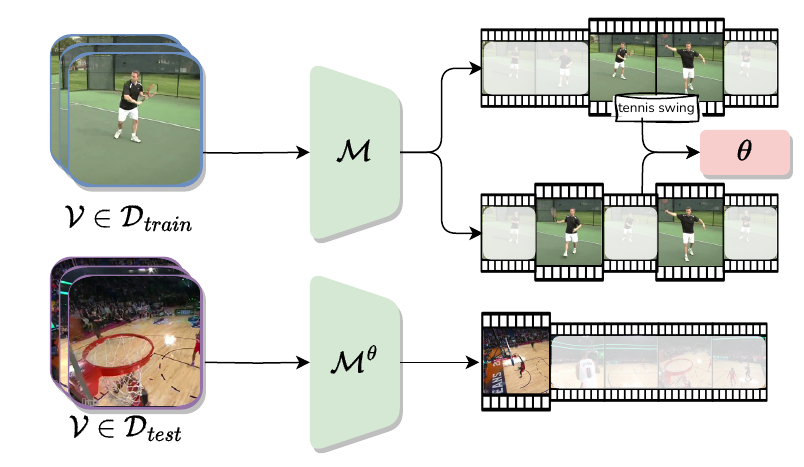}}}
    \vspace{10pt}
    \subfloat[Our proposal]{{\includegraphics[width=0.9\linewidth]{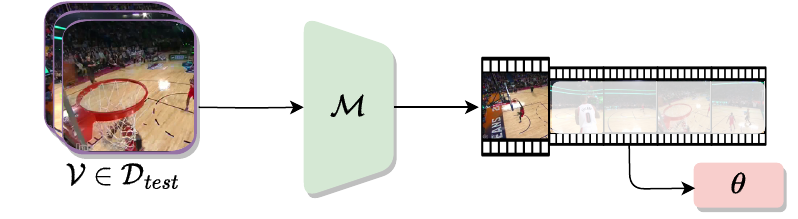} }}
    \caption{\textbf{Task setup.} Previous approaches tackling ZS-TAL (a) train the \inlineColorbox{ModelGreen}{model} on labelled data and test it in-domain. Due to lack of out-of-distribution generalization, we propose to update the \inlineColorbox{ParamRed}{parameters} at test-time on a stream of unlabelled videos  without prior supervised training  (b).}
    \label{fig:teaser}
    \vspace{-4mm}
\end{figure}

Zero-shot Temporal Action Localization (ZS-TAL) aims to locate and recognize actions in any video sequence, enabling the recognition of classes unseen during training.
Large-scale Vision and Language models (VLMs)~\cite{radford2021learning,flamingo,blip2,yu2022coca,xu2021videoclip} are renowned for the exceptional generalization capability derived from their extensive pre-training on web-scale image-text datasets, outperforming traditional image classification methods~\cite{goyal2023finetune,zara2023unreasonable,li2022grounded}.
When applied to the video domain, however, VLMs typically require fine-tuning to account for the image-video structural domain shift~\cite{xu2021videoclip, pmlr-v162-li22n,wu2023bidirectional}.

Recent methods exploiting VLMs for ZS-TAL also abide by this limitation, requiring training data to learn the video domain and localize unseen actions at test time~\cite{Ju2021PromptingVM,stale,Yan_2023_ICCV} (see Fig.~\ref{fig:teaser}(a)).

While model fine-tuning has the clear objective of learning video representations, which allows to effectively localize actions in the untrimmed videos, it also assumes the availability of a large annotated data collection. In certain applications, however, such datasets may be unavailable. Furthermore, fine-tuning inherently carries the downside of producing models with decreased out-of-domain generalization capabilities~\cite{zhou2022conditional}. 
This latter issue is especially severe for ZS-TAL.
A preliminary investigation (see Sec.~\ref{sec:pre}) demonstrates that state-of-the-art ZS-TAL evaluated in a cross-domain setting suffer from a dramatic drop in performance. This clearly raises concerns regarding the adaptability and robustness of existing ZS-TAL approaches, expecially in real-world scenarios where training data is inaccessible due to privacy concerns or when a major data distribution shift occurs over time.

Motivated by these observations, in this work we propose to investigate the problem of ZS-TAL under a novel perspective, featuring the relevant scenario where training data is inaccessible. Even with the aid of powerful VLMs, locating and recognizing actions without training is undoubtedly challenging, since videos carry additional complexities with respect to images, induced by the scene clutter and the difficulty of modelling the temporal dynamics~\cite{momeni2023verbs}. Nonetheless, we argue that, even in the absence of training data, videos made available at inference time can be exploited as a rich source of information for temporal action localization.

Inspired by recent works on Test-Time Adaptation (TTA)~\cite{wang2021tent, sun19ttt}, we propose \textbf{T3AL}, standing for \textbf{T}est \textbf{T}ime adaptation for \textbf{T}emporal \textbf{A}ction \textbf{L}ocalization. \methodname~utilizes a pre-trained VLM not fine-tuned on training data, as opposed to previous works~\cite{Ju2021PromptingVM,stale,Yan_2023_ICCV}. Our solution instead, undergoes direct adaptation when a video sequence is made available during inference (Fig.~\ref{fig:teaser}(b)). \methodname~unfolds in three key steps. First, we compute video-level pseudo-labels corresponding to action categories by aggregating semantic information extracted from each frame by the VLM image encoder. Next, a first solution to temporal action localization is produced, based on the results from video pseudo-labels by employing a novel procedure inspired by self-supervised learning~\cite{byol}. The computed video action proposals are then refined with frame-level textual descriptions extracted from a state-of-the-art captioning model.
We evaluate \methodname~on two publicly available benchmarks, \textit{i.e.} THUMOS14~\cite{IDREES20171} and ActivityNet-v1.3~\cite{7298698}, achieving a relative improvement of $+6.3\%$ and $+13.5\%$ when compared to a naive application of VLMs for TAL without TTA. Through several oracle experiments, we also demonstrate the existence of a potential space for future improvements. This suggests that test-time ZS-TAL is a viable approach for locating and recognising actions in arbitrary video sequences. We hope that these findings will inspire future research in this area.

\noindent Our contributions can be summarized as follows: 
\begin{itemize}
    \item We address ZS-TAL in a new practical scenario where training data is unavailable. We demonstrate that this is a challenging problem as state-of-the-art methods for the task poorly generalize without training.
    \item We propose \methodname, the first method that tackles ZS-TAL without training data by leveraging a pre-trained VLM. \methodname~benefits from an effective TTA strategy and from external knowledge derived from generated captions.
    \item We empirically demonstrate that adapting on an unlabeled stream of data is a viable solution to the out-of-distribution issue of current training-based approaches for ZS-TAL.
\end{itemize}
\section{Related work}
Our work is closely related to existing literature in Zero-Shot Temporal Action Localization and Test-Time Adaptation (TTA), which we briefly review in the current section.

\textbf{Zero-Shot Temporal Action Localization.}
Temporal Action Localization (TAL) methods jointly perform action localization and recognition. 
Existing works either tackle the problems sequentially, \ie, two-stage~\cite{Zhao_2017_ICCV, Xu_2017_ICCV, Chao_2018_CVPR, Ju2021PromptingVM}, or concurrently, \ie, one-stage~\cite{sstad_buch_bmvc17, lin_acm, stale, Yan_2023_ICCV}. Two-stage methods first identify class-agnostic region proposals and then classify each region. One-stage methods perform action localization and classification simultaneously.
Traditionally, both one-stage and two-stage approaches work in closed-set scenarios, where train and test data share the same action categories.
Recently, EffPrompt~\cite{Ju2021PromptingVM} introduced the novel ZS-TAL setup, removing the above premise and isolating action categories between training and testing.
To address the novel setup, EffPrompt employs a two-stage architecture to generate action proposals with an off-the-shelf detector~\cite{Lin_2021_CVPR} and then classifies action proposals with CLIP~\cite{radford2021learning}. 
Differently, STALE~\cite{stale} proposes to train a CLIP-based proposal-free model using two concurrent streams for localization and classification. The localization branch focuses on learning a class-agnostic representation masking, while the classification stream aligns the masked features with the text embeddings of the respective class, contributing to the final classifier output. More recently, UnLoc~\cite{Yan_2023_ICCV} extracts joint features for video-text pairs with CLIP and feeds them into a dedicated fusion module. A feature pyramid architecture then takes these refined outputs and establishes hierarchical connections, predicting per-frame relevance scores and start/end time displacements. 

Although effective, existing ZS-TAL methods require learning a model on a training set, leading to several inherent limitations. As discussed above, these limitations encompass challenges in generalization to out-of-domain data, high computational requirements and reliance on availability of annotated data. We tackle a more practical yet challenging setup for ZS-TAL where the training set is inaccessible. Our method falls into the one-stage category, addressing the challenges of ZS-TAL in an integrated manner.

\textbf{Test-Time Adaptation.}
In TTA, a model pre-trained on a training dataset must adapt to an unknown test distribution expressed as an unlabelled data stream~\cite{wang2021tent, sun19ttt}. Several works propose TTA methods for image classification. For instance, TENT~\cite{wang2021tent} adapts a pre-trained network at test-time by minimizing the entropy of the batch-wise prediction probability distributions. Similarly, MEMO~\cite{memo} reduces the entropy of the marginal distribution across augmentations, overcoming the necessity for multiple samples. Recent works explore these concepts for large-scale VLMs~\cite{tpt,ma2023swapprompt,samadh2023align}. TPT~\cite{tpt} adapts CLIP by learning textual context vectors via entropy minimization. SwapPrompt~\cite{ma2023swapprompt} employs a swapping mechanism between the online prompt and its historical moving average to improve and stabilize adaptation. 
PromptAlign~\cite{samadh2023align} fine-tunes multi-modal prompts at test-time by aligning the distribution statistics obtained from multiple augmented views of a single test image with the training data distribution statistics.

While all these methods tackle TTA in the image domain, videos still remains largely unexplored. A notable exception is ViTTA~\cite{lin2023video}, which performs test-time adaptation for video action recognition, handling the distribution shifts and aligning test and pre-computed train statistics online. Another contribution in this area is RNA$^{++}$~\cite{egotta} which proposes a TTA approach to address the problem of domain shift in egocentric action recognition. This is particularly relevant when unsupervised domain adaptation approaches, while effective, are impractical due to the unavailability of data from the target distribution. 
Our research differs significantly from~\cite{lin2023video, egotta}, as we adapt models not specifically designed to resolve the TAL task. This necessitates the model to deduce a previously unencountered task.
\section{Cross-dataset generalization analysis}\label{sec:pre}
We propose a preliminary experiment to motivate further the novel research direction proposed in this work. The goal is to test the generalization capability of state-of-the-art methods for ZS-TAL, testing their action localization performance in a cross-dataset setting. 

Specifically, we consider two state-of-the-art methods, EffPrompt~\cite{Ju2021PromptingVM} and STALE~\cite{stale}\footnote{No public code implementation was available for UnLoc~\cite{Yan_2023_ICCV} at the time of the submission.}.
For EffPrompt~\cite{Ju2021PromptingVM}, we use their off-the-shelf detector~\cite{Lin_2021_CVPR} and their action recognition model trained on HMDB51~\cite{6126543}.
For STALE~\cite{stale}, we use their ZS-TAL model trained on ActivityNet-v1.3.
We conduct both experiments using THUMOS14 as a target dataset.
Note that our \textit{cross-domain} protocol evaluates these models, which are trained on more challenging and diverse datasets (\ie, ActivityNet-v1.3, 200 classes; HMDB51, 51 classes), on a simpler data collection (\ie, THUMOS14, 20 classes).
For reference, we also report results obtained in an \textit{in-domain} setting, \ie where both methods are trained and tested on THUMOS14.

\begin{figure}[!t]
    \centering
    \includegraphics[width=1.0\linewidth]{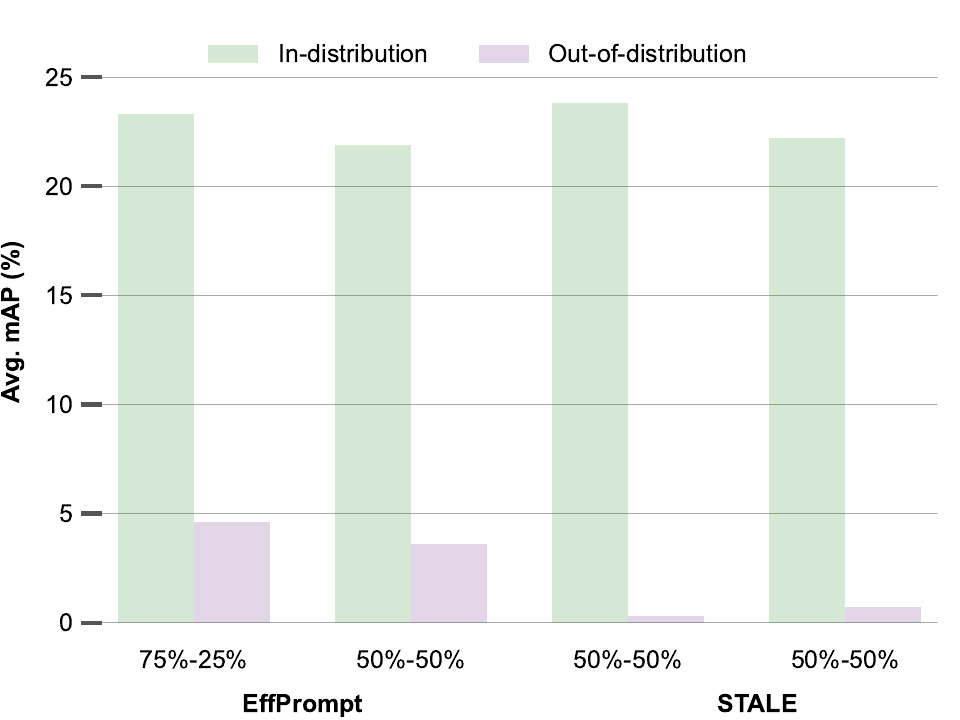}
    \caption{\textbf{Cross-dataset generalization.} We show the average mAP, computed at IoU thresholds of [$0.3$:$0.1$:$0.7$], for EffPrompt and STALE trained and tested on \inlineColorbox{ModelGreen}{THUMOS14}, and trained on a \colorbox{FunctionPurple}{different dataset} and tested on THUMOS14. We report results for the 75:25 (75\% seen classes) and 50:50 (50\% seen classes) evaluation settings.}
    \label{fig:cross_dataset}
\end{figure}

The results of our preliminary investigation show that EffPrompt and STALE do not generalize on out-of-distribution samples, despite i) being trained to improve their zero-shot capabilities (i.e., their ability to recognize unseen classes) and ii) the usage of prior knowledge encoded in VLMs. Fig.~\ref{fig:cross_dataset} shows the comparison between the performance of the two methods in a \textit{in-domain} and a \textit{cross-domain} setting. The plot shows a drastic drop (\ie, higher than 15\% mAP in both settings) in the performance of both methods when the test dataset differs with respect to the one used for model fine-tuning.
We associate this behavior to the fact that a perturbation of the weights of the VLM boosts \textit{in-domain} prediction ability, but hinders \textit{out-domain} generalization. 
Motivated by these experimental findings, we devise a method to achieve robust performance on different datasets without the need for annotated training data.

\begin{figure*}[!t]
    \centering
    \includegraphics[width=1.\linewidth]{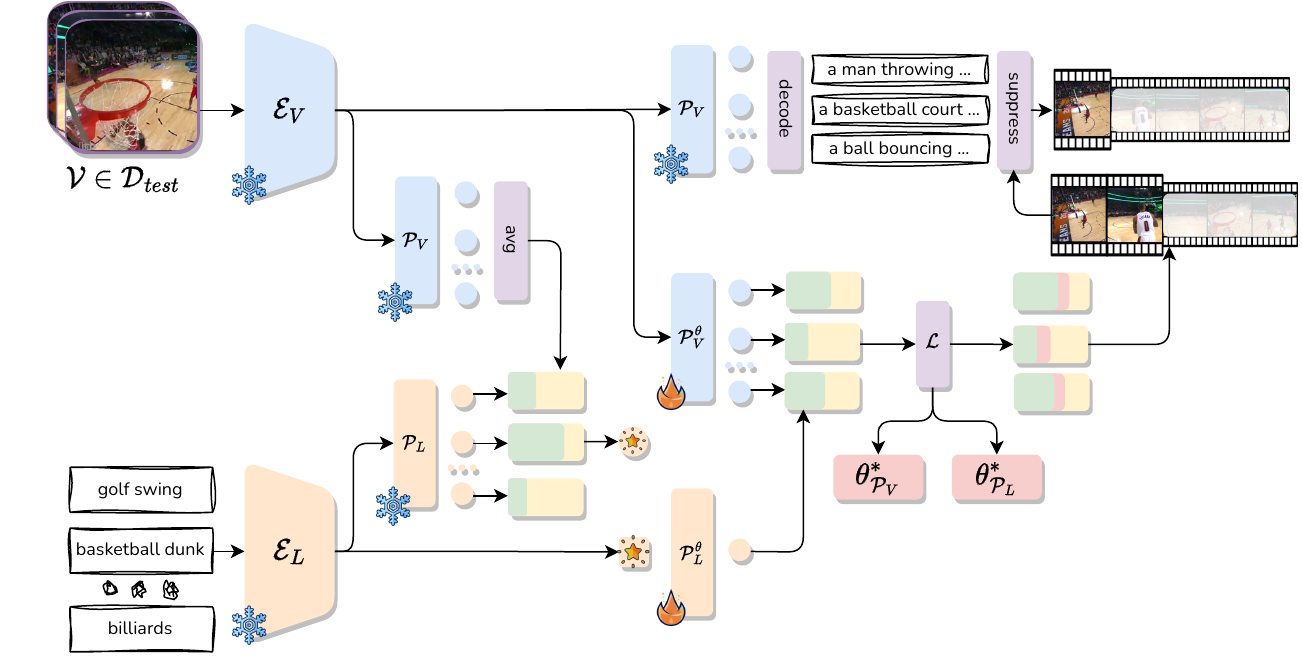}
    \caption{\textbf{Overview of the proposed method.} \methodname~addresses the task of ZS-TAL by only learning at test-time on unlabelled data. We first compare the \inlineColorbox{FunctionPurple}{average} \inlineColorbox{VisionBlue}{visual frames} with the \inlineColorbox{LanguageOrange}{textual class names} to identify the video pseudo-label \raisebox{-1.3mm}{\includegraphics[height=4.3mm]{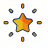}}. We then refine the \inlineColorbox{VisionBlue}{visual frames}-\inlineColorbox{LanguageOrange}{video pseudo-label} scores with self-supervision. Last, we exploit the \inlineColorbox{FunctionPurple}{decoder} of a captioning model (\ie, CoCa~\cite{yu2022coca}) to generate captions and perform text-guided \inlineColorbox{FunctionPurple}{region suppression}. We only \raisebox{-1.2mm}{\includegraphics[height=4.0mm]{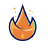}} fine-tune the \inlineColorbox{VisionBlue}{vision} and \inlineColorbox{LanguageOrange}{language} projectors, while keeping the encoders \raisebox{-1.0mm}{\includegraphics[height=3.6mm]{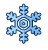}} frozen. 
    Once the prediction is obtained, the optimized \inlineColorbox{ParamRed}{parameters} $\theta_{\encoderprojector_V}^\ast$ and $\theta_{\encoderprojector_L}^\ast$ are re-initialized to the ones of the pre-trained model.}
    \label{fig:method}
\end{figure*}

\section{T$^3$AL: Test Time Adaptation for Temporal Action Localization}\label{sec:method}
In this section we first define the problem, then we detail the main steps of \methodname: video-level pseudo-labelling, self-supervised prediction refinement and text-guided region suppression. 

\subsection{Problem definition}
A ZS-TAL algorithm aims to identify and classify actions in untrimmed videos. For each detected temporal region, the model predicts the class and indicate when it starts and ends. 
While the set of classes is given, they differ from the categories seen by the model at training.
Existing literature addressing ZS-TAL ~\cite{Ju2021PromptingVM,stale,Yan_2023_ICCV} always involves a labelled training set $\trainingset$ and a test set $\testset$, with two disjoint sets of action classes. Yet, as shown in Sec.~\ref{sec:pre}, state-of-the-art methods greatly rely on $\trainingset$, resulting in poor generalization if the two datasets are drawn from different distributions. In this paper, we advocate for the need to investigate a different scenario for ZS-TAL—relevant for practical applications—where \textit{in-domain} training data is unavailable. 

Given a video $\video$, our goal is to localize regions and assign corresponding actions from a set of classes $\actionclassset$, without having access to $\trainingset$. The $\numpredictions$ predicted action proposals are defined as $\{\left(y_i, t_i\right)\}_{i=1}^\numpredictions$, where ${y_i}\in\actionclassset$ is the class and $t_i\in\mathbb{R}^2$ is the start/end time displacement of each action.

\methodname~is built on top of a pre-trained VLM model $\model$, consisting of a vision encoder $\visionencoder$ and a language encoder $\textencoder$, as shown in Fig.~\ref{fig:method}. \methodname~directly addresses ZS-TAL at inference time only exploiting a single test video at a time. First, the frame-level representations extracted from the visual frames with $\visionencoder$ are averaged and compared with the class textual representations to identify the video pseudo-label. Then, the scores of the visual frames are computed and refined adapting $\model$ at test-time with self-supervision. Finally, we exploit a captioning model to generate captions and perform an additional suppression step. 
\methodname~is applied on a sample basis. Once the prediction for one sample is made, the optimized parameters of $\model$ are reverted to the original initialization.

\subsection{Video-level pseudo-labelling}
At test-time the only accessible information is the unlabelled sample $\video=\{x_i\}_{i=1}^\numframes$ and the set of action categories $\actionclassset$.
However, we can mitigate such lack of supervision by leveraging the knowledge already encoded in $\mathcal{M}$, as VLMs have demonstrated strong zero-shot capabilities in a wide range of classification tasks. First, we compute a compact representation from $\video$ as the average of its $\numframes$ frames' latent representations extracted with the vision encoder:
\begin{equation}
\bar{\video} = \frac{1}{\numframes} \sum\limits_{i=1}^{\numframes} \visionencoder\left(x_i\right)
\end{equation}
In this compact representation the noise coming from non-informative frames present in the video is mitigated. Thus, we can exploit $\bar{\video}$ to compute a pseudo-label $y^\ast$\footnote{In Fig~\ref{fig:method}, the pseudo-label is encoded as a star.} for the whole video $\video$, selecting the label with the maximum cosine similarity: 
\begin{equation}
y^\ast = \argmax_{y\in\actionclassset}\cos\left(\bar{\video}, \textencoder\left({y}\right)\right)
\end{equation}
where $\cos\left(\cdot,\cdot\right)$ indicates the cosine similarity.  We propose to use $y^\ast$ to guide the localization process, providing a foundation for more temporally fine-grained predictions.

\subsection{Self-supervised prediction refinement}\label{selfsupscore}
The objective of the second step of our method is to refine this coarse-grained video prediction to effectively localize regions in $\video$ where the action of class $y^\ast$ occurs. The video comprises frames that capture the pseudo-label $y^\ast$, alongside frames that do not exhibit any correspondence with it. The model $\mathcal{M}$ can easily classify these frames, but struggles on those lying in-between the two extremes as they involve visual cues semantically linked to $y^\ast$ while neglecting the actual execution of the action. Following this intuition, we propose to utilize those frames on which the model $\mathcal{M}$ is confident to filter out the noise in the prediction of the more ambiguous ones. Therefore, we compute the semantic closeness of every frame in the video to $y^\ast$, assigning the score:
\begin{equation}
    \score_i=\cos(\visionencoder(\image_i), \textencoder\left( y^\ast \right)).
\end{equation}
and for each frame in $\video$ we denote the corresponding feature representation as $z_i=\visionencoder(\image_i)$. 

Frames with higher scores are more likely to be associated with the foreground, \ie, actions present in the video, while frames with lower scores are more likely to correspond to the background, representing non-action-related content. Building on this consideration, we aim to strategically leverage these frames with a self-supervised objective to refine the initial predictions. Specifically, we form a set of positive samples $\positiveset$ from features with higher scores, and a set of negative samples $\negativeset$ from features with lower scores: 

\begin{equation}
\positiveset=\{\left(z_i^+, \score_i^+\right)\}_{i=1}^K,\quad \negativeset=\{\left(z_i^-, \score_i^-\right)\}_{i=1}^K
\end{equation}
For both sets, the $K$ features to be selected are distributed over the temporal dimension with a slight perturbation governed by a random noise $\epsilon$, avoiding concentration and ensuring diversity in the selection.

Our self-supervised objective can be formulated as: 
\begin{equation}
\left(\theta_{\encoderprojector_V}^\ast, \theta_{\encoderprojector_L}^\ast, \tau^\ast\right)= \argmin_{\theta_{\encoderprojector}, \tau}\mathcal{L}
\end{equation}
where we only adapt the parameters of the two projections $\theta_{\encoderprojector}=\left(\theta_{\encoderprojector_V}, \theta_{\encoderprojector_L}\right)$, and the temperature parameter $\tau$. 

The loss $\mathcal{L}$ can be further decomposed into two, with one taking as input the visual representations of the frames and the other taking as input the scores:
\begin{equation}
\mathcal{L} = \mathcal{L}_z + \mathcal{L}_s,  
\end{equation}
where $\mathcal{L}_z$ is the \lossx~and $\mathcal{L}_s$ is the \losss.
The \lossx~is exclusively applied to the positive set $\positiveset$ to bring positive frames closer in the embedding space. 
Differently, we refrain from applying the same objective to negative samples, as the background frames of the video carry much diverse information with various non-action-related visual cues. Without any guarantees of a shared semantics, we should not force the representations of these frames to be represented closer in space.

To address this, we adopt the BYOL~\cite{byol} loss, commonly utilized in self-supervised learning, particularly in scenarios lacking negative examples. 

This \lossx~ enforces both visual and semantic closeness among potentially repeated instances of the same action within the video. 
Additionally, we assume that semantic knowledge is a continuous function, \ie, the information in a frame at time $t$ is likely to be similar to that in adjacent frames, incorporating them into the set of positive candidates.

While the BYOL loss requires augmented views of a sample, we exploit the natural temporal dimension of videos to obtain multiple views for free. Following these observations, all aforementioned positive samples can be seen as augmented views and used to compute the loss as:

\begin{equation}
\mathcal{L}_z = 2 - 2 \cdot \frac{<z^+_i, z^+_j>}{\lVert z^+_i\lVert_2 \cdot \lVert z_j^+\lVert_2}
\end{equation}
where $z^+_i$ and $z^+_j$ are randomly sampled at each step. 

The \losss~is applied to both $\positiveset$ and $\negativeset$ and aims to push the scores of positive samples closer to $1$ and the negative ones to $0$, promoting their separation. It is again implemented as a BYOL loss (this component is ablated in Sec.~\ref{sec:abl}). Specifically, we define the prediction vector as the concatenation of positive ad negative scores:

\begin{equation}
\mathbf{\score}=\texttt{concat}\left(\{\score_i^+\}_{i=1}^K,\{\score_i^-\}_{i=1}^K\right)\in\mathbb{R}^{2K}
\end{equation}
and the binary target vector accordingly:
\begin{equation}
\mathbf{b} = \begin{bmatrix} 1_K \\ 0_K \end{bmatrix} \in\mathbb{R}^{2K}
\end{equation}
Then, the loss is computed as: 
\begin{equation}
 \mathcal{L}_s = 2 - 2 \cdot \frac{<\mathbf{\score},\mathbf{b}>}{\lVert \mathbf{\score}\lVert_2 \cdot \lVert \mathbf{b}\lVert_2}. 
\end{equation}

At each step in the test-time adaptation, we recompute $\positiveset$ and $\negativeset$. After $T$ steps, the adapted model assigns a final score to each frame in $\video$. We compute a moving average of these scores to further enhance temporal consistency. We then obtain temporal action proposals, \ie, start/end time displacements $\{{\hat{t}_i}\}_{i=1}^{{\hat{\numpredictions}}}$, by filtering with a threshold $\gamma$. 
Instead of using a fixed threshold, we set $\gamma$ as the average value of the scores along the whole video. After filtering, we group consecutive frames into region proposals and update each region label $\hat{y}_i$ with the class that maximizes the cosine similarity of the region-level representation, defined as the average of its frames.

\subsection{Text-guided region suppression}\label{textguided} 

The last step aims to reduce potential incorrectly predicted action proposals. To achieve this, we utilize the semantic guidance of a existing captioning model, to contribute in identifying semantic variations from the textual modality.

First, all frames belonging to the selected action proposals are captioned. Then, we feed the obtained captions to the language encoder $\textencoder$, averaging the textual representations obtained within each temporal proposal $\hat{t}_i$ to get a region-level representation $d_i$ that carries semantic information. We establish a rejection criteria by calculating the pairwise cosine similarity among all these representations. To this aim, we define the matrix of pairwise cosine similarities as: 
\begin{equation}
    \mathbf{D}=\left[d_{ij}\right],\quad d_{ij}=\cos\left(d_i, d_j\right)
\end{equation}
Then, we binarize it at a threshold $\beta$ and obtain $\hat{\mathbf{D}}$. Summing up column-wise the elements in this binary mask, we obtain a score vector $\mathbf{d}=\hat{\mathbf{D}}\diag \left(\text{I}_{\hat{M}}\right)\in\mathbb{R}^{\hat{M}}$ that measures the similarity of each action proposal with the others.

At last, a proposal $\hat{t}_i$ is suppressed if its associated entry in $\mathbf{d}$ is below a threshold $\alpha$, \ie, its associated textual representation is insufficiently close to the others. As a result, we obtain $\{\left(y_i, t_i\right)\}_{i=1}^\numpredictions$, with $M\le \hat{M}$. After making a prediction on $\video$, the optimized parameters $\left(\theta_{\encoderprojector_V}^\ast, \theta_{\encoderprojector_L}^\ast, \tau^\ast\right)$ are discarded and re-initialized with the original ones.

We adopt the CoCa~\cite{yu2022coca} model. This model follows a dual encoder architecture and includes an extra text decoder, and is trained with a contrastive loss and a captioning loss. This model is particularly convenient for our approach since it allows us to adopt a single model to perform action classification as well as proposal generation and suppression, via its the textual/visual encoder and the underlined captioner.

\section{Experiments}
\noindent\textbf{Datasets and settings.} We conduct experiments with two popular untrimmed video datasets, \ie, ActivityNet-v1.3~\cite{7298698} and THUMOS14~\cite{IDREES20171}. ActivityNet-v1.3 contains 19,994 videos describing 200 action classes, while THUMOS14 has 413 videos from 20 categories. Following~\cite{Ju2021PromptingVM}, we validate our approach by dividing dataset classes into training and testing. We consider a 50\%-50\% split and a 75\%-25\% split. To guarantee statistical significance, we repeat class sampling ten times per split, reporting their average.

\noindent\textbf{Metrics.} We report the mean Average Precision (mAP) computed at different temporal IoU thresholds. Following prior work~\cite{Ju2021PromptingVM, stale}, our tables include mAP at IoU thresholds of [$0.3$:$0.1$:$0.7$] for the THUMOS14 dataset and [$0.5$:$0.05$:$0.95$] for the ActivityNet-v1.3 dataset.

\noindent\textbf{Implementation details.}
We extract RGB frames maintaining the original frame rate and resizing them to a resolution of $224\times224$. Class names are augmented with the prompt ``\texttt{a video of action \{CLS\}}", for both \methodname~and the baselines defined in Sec.~\ref{comparison}. We use CoCa (ViT-L/14) with the implementation of~\cite{ilharco_gabriel_2021_5143773}. We adapt it for a maximum of $T=50$ steps on THUMOS14 and $T=25$ on ActivityNet-v1.3, with early stopping on the individual sample: if the loss does not diminish after 5 consecutive steps, the adaptation process is halted and we proceed to the final prediction. We use Adam optimizer with a learning rate of $1e^{-5}$, and set $\alpha=0.5$, $\beta=0.75$, and $K=4/20$ for THUMOS14 and ActivityNet-v1.3. We empirically observe that subtracting $y^\ast$ from the visual features improves the performance on THUMOS14, augmenting the discrimination between features belonging to foreground and background of the action. We observe the opposite on ActivityNet-v1.3 and attribute this behaviour to the different average video length. We therefore remove the background information only for THUMOS14. All the experiments are conducted using one NVIDIA V100 GPU in floating point precision.

\subsection{Comparative results}\label{comparison}
\begin{table}[t]
\small
\centering
\resizebox{\columnwidth}{!}{
\begin{tabular}{lccccc|c}
\toprule
\textbf{Method} & \multicolumn{6}{c}{\textbf{mAP (\%) $\uparrow$}} \\
                & 0.3 & 0.4 & 0.5 & 0.6 & 0.7 & \textbf{Avg.}  \\ 
\midrule
CLIP$_{32}$~\cite{radford2021learning} & 7.2 & 4.1 & 2.3 & 1.1 & 0.5 & 3.0 \\
CLIP$_{16}$~\cite{radford2021learning} & 7.5 & 4.2 & 2.2 & 1.1 & 0.6 & 3.1 \\
CoCa~\cite{yu2022coca}                 & 8.4 & 4.7 & 2.5 & 1.2 & 0.6 & 3.5 \\
\midrule
\methodname$_{T=0}$                    & 11.4 & 6.8 & 3.5 & 1.7 & 0.6 & 4.8 \\
\rowcolor{ModelGreen}
\methodname                            & 20.7 & 14.3 & 8.9 & 5.3 & 2.7 & 10.4 \\
\midrule
\rowcolor{FunctionPurple}
CLIP$_{16}$ w\slash~Detector~\cite{Ju2021PromptingVM,stale}     & 27.2 & 21.3 & 15.3 & 9.7 & 4.8 & 15.7 \\
\rowcolor{FunctionPurple}
EffPrompt~\cite{Ju2021PromptingVM}     & 37.2 & 29.6 & 21.6 & 14.0 & 7.2 & 21.9 \\
\rowcolor{FunctionPurple}
STALE~\cite{stale}                     & 38.3 & 30.7 & 21.2 & 13.8 & 7.0 & 22.2 \\
\bottomrule
\end{tabular}
}
\caption{\textbf{Results on THUMOS14 (50\%-50\%).} Green is \inlineColorbox{ModelGreen}{our method}, purple indicates \inlineColorbox{FunctionPurple}{training-based} approaches.}
\label{thumos50}
\end{table}
\begin{table}[t]
\small
\centering
\resizebox{1.\columnwidth}{!}{
\begin{tabular}{lccccc|c}
\toprule
\textbf{Method} & \multicolumn{6}{c}{\textbf{mAP (\%) $\uparrow$}} \\
                & 0.3 & 0.4 & 0.5 & 0.6 & 0.7 & \textbf{Avg.}  \\ 
\midrule
CLIP$_{32}$~\cite{radford2021learning} & 5.5 & 3.3 & 1.9 & 0.9 & 0.4 & 2.4 \\
CLIP$_{16}$~\cite{radford2021learning} & 6.9 & 3.8 & 2.1 & 1.1 & 0.6 & 2.9 \\
CoCa~\cite{yu2022coca}                 & 7.8 & 4.6 & 2.5 & 1.3 & 0.6 & 3.4 \\
\midrule
\methodname$_{T=0}$                    & 11.1 & 6.5 & 3.2 & 1.5 & 0.6 & 4.6 \\
\rowcolor{ModelGreen}
\methodname                            & 19.2 & 12.7 & 7.4 & 4.4 & 2.2 & 9.2 \\
\midrule
\rowcolor{FunctionPurple}
CLIP$_{16}$ w\slash~Detector~\cite{Ju2021PromptingVM,stale}                       & 33.0 & 25.5 & 18.3 & 11.6 & 5.7 & 18.8 \\
\rowcolor{FunctionPurple}
EffPrompt~\cite{Ju2021PromptingVM}     & 39.7 & 31.6 & 23.0 & 14.9 & 7.5 & 23.3 \\
\rowcolor{FunctionPurple}
STALE~\cite{stale}                     & 40.5 & 32.3 & 23.5 & 15.3 & 7.6 & 23.8  \\
\bottomrule
\end{tabular}
}
\caption{\textbf{Results on THUMOS14 (75\%-25\%).} Green is \inlineColorbox{ModelGreen}{our method}, purple indicates \inlineColorbox{FunctionPurple}{training-based} approaches.}
\label{thumos75}
\end{table}

\begin{table}[t]
\small
\centering
\resizebox{0.85\columnwidth}{!}{
\begin{tabular}{lccc|c}
\toprule
\textbf{Method} & \multicolumn{4}{c}{\textbf{mAP (\%) $\uparrow$}} \\
                & 0.50 & 0.75 & 0.95 & \textbf{Avg.}  \\ 
\midrule
CLIP$_{32}$~\cite{radford2021learning} & 0.4 & 0.2 & 0.0 & 0.2 \\
CLIP$_{16}$~\cite{radford2021learning} & 0.8 & 0.3 & 0.0 & 0.3 \\
CoCa~\cite{yu2022coca}                 & 2.3 & 1.0 & 0.2 & 1.1 \\
\midrule
\methodname$_{T=0}$                    & 24.2 & 13.0 & 2.8 & 13.3 \\
\rowcolor{ModelGreen}
\methodname                            & 25.8 & 13.9 & 3.1 & 14.3 \\
\midrule
\rowcolor{FunctionPurple}
CLIP$_{16}$ w\slash~Detector~\cite{Ju2021PromptingVM,stale}  & 28.0 & 16.4 & 1.2 & 16.0 \\
\rowcolor{FunctionPurple}
EffPrompt~\cite{Ju2021PromptingVM}     & 32.0 & 19.3 & 2.9 & 19.6 \\
\rowcolor{FunctionPurple}
STALE~\cite{stale}                     & 32.1 & 20.7 & 5.9 & 20.5 \\
\rowcolor{FunctionPurple}
UnLoc~\cite{Yan_2023_ICCV}             & 43.7 & -  & -  & -  \\
\bottomrule
\end{tabular}
}
\caption{\textbf{Results on ActivityNet-v1.3 (50\%-50\%).} Green is \inlineColorbox{ModelGreen}{our method}, purple indicates \inlineColorbox{FunctionPurple}{training-based} approaches.}
\label{activitynet50}
\end{table}

\begin{table}[t]
\small
\centering
\resizebox{0.85\columnwidth}{!}{
\begin{tabular}{lccc|c}
\toprule
\textbf{Method} & \multicolumn{4}{c}{\textbf{mAP (\%) $\uparrow$}} \\
                & 0.50 & 0.75 & 0.95 & \textbf{Avg.}  \\ 
\midrule
CLIP$_{32}$~\cite{radford2021learning} & 0.4 & 0.1 & 0.0 & 0.2 \\
CLIP$_{16}$~\cite{radford2021learning} & 0.9 & 0.3 & 0.1 & 0.4 \\
CoCa~\cite{yu2022coca}                 & 3.1 & 1.3 & 0.3 & 1.6 \\
\midrule
\methodname$_{T=0}$                    & 26.1 & 13.9 & 2.9 & 14.3 \\
\rowcolor{ModelGreen}
\methodname                            & 28.1 &	14.9 &3.3 &	15.4 \\
\midrule
\rowcolor{FunctionPurple}
CLIP$_{16}$ w\slash~Detector~\cite{Ju2021PromptingVM,stale}   & 35.6 &  20.4 & 2.1 & 20.2 \\
\rowcolor{FunctionPurple}
EffPrompt~\cite{Ju2021PromptingVM}     & 37.6 & 22.9 & 3.8 & 23.1 \\
\rowcolor{FunctionPurple}
STALE~\cite{stale}                     & 38.2 & 25.2 & 6.0 & 24.9 \\
\rowcolor{FunctionPurple}
UnLoc~\cite{Yan_2023_ICCV}             & 48.8 & -  & - & -  \\
\bottomrule
\end{tabular}
}
\caption{\textbf{Results on ActivityNet-v1.3 (75\%-25\%).} Green is \inlineColorbox{ModelGreen}{our method}, purple indicates \inlineColorbox{FunctionPurple}{training-based} approaches.
\vspace{-10pt}}
\label{activitynet75}
\end{table}
As we are unaware of methods designed for ZS-TAL that abstain from training on labeled data, we propose three baselines on top of pre-trained VLMs: CLIP (ViT-B/32), CLIP (ViT-B/16)~\cite{radford2021learning}, and CoCa (ViT-L/14)~\cite{yu2022coca}. In the following, we will refer to these as CLIP$_{32}$, CLIP$_{16}$, and CoCa. For each of the three, the naive approach for TAL consists of independently classifying the video frames. We evaluate the cosine similarity between frame-level representations and textual descriptions generated via prompting on the class names. We convert their image-text cosine similarities into probabilities with the softmax operator. To recognize frames as actions or background, we binarize their probabilities for the predicted class. For this, we apply a threshold of $0.8$. 
Following previous work~\cite{Ju2021PromptingVM, stale}, we report a two-stage baseline (which we call CLIP$_{16}$ w\slash~Detector) consisting of a pre-trained proposal detector~\cite{Lin_2021_CVPR} and CLIP as the second-stage proposal classifier. 
As an additional baseline, we present our method with $T=0$ steps of adaptation, denoted as \methodname$_{T=0}$. Tab.~\ref{thumos50} and Tab.~\ref{thumos75} show results on THUMOS14 for the 50\%-seen 50\%-unseen and 75\%-seen 25\%-unseen splits; Tab.~\ref{activitynet50} and Tab.~\ref{activitynet75} present the same class splits on ActivityNet-v1.3. For the ease of readers, we also report results on state-of-the-art models achieved through training.~\cite{Ju2021PromptingVM, stale, Yan_2023_ICCV}.
All the tables suggest that a naive application of VLMs is insufficient for the ZS-TAL task. We further show that a simple use of the video-level pseudo-labelling, as detailed in Sec.~\ref{sec:method}, can considerably improve results of these baselines. \methodname$_{T=0}$ without TTA achieves an average improvement of +1.2\% mAP on THUMOS14, and of +12.4\% mAP on ActivityNet-v1.3. With test-time learning we improve further, with an extra gain of +1.0\% and +5.1\% mAP on ActivityNet-v1.3 and THUMOS14. Refer to the \textit{Supplementary Material} for additional results.

\subsection{Ablation}\label{sec:abl}

We thoroughly ablate \methodname~on the THUMOS14 dataset to validate the effectiveness of our design choices. First, we analyze the learning objective functions, followed by the selected fine-tuned parameters and the final suppression step. Furthermore, we conduct oracle experiments to showcase the potential of our approach in addressing this challenging scenario.
When not state otherwise, we report results for the 50\%-seen 50\%-unseen split, averaged across the 10 splits to guarantee statistical significance.

\noindent{\textbf{Learning objective.}} In Tab.~\ref{ablation_loss} we analyze the learning objective and the selected parameters for adaptation. We compare different configurations of the loss defined in Sec.~\ref{selfsupscore} and a binary cross entropy (BCE) loss on the same input of $\mathcal{L}_z$. Notably, incorporating the loss on the representations (see Row 3-4) improves performance compared to solely utilizing the loss on the scores (see Row 1-2). When the \lossx~is added to the \losss~and we adapt both the vision and language projection layers, we observe an improvement up to +2.0\% mAP. 
\begin{table}[t]
\small
\centering
\resizebox{0.9\columnwidth}{!}{
\begin{tabular}{lll|ccccc|c}
\toprule
\multicolumn{3}{l}{\textbf{Configuration}}                                  & \multicolumn{6}{c}{\textbf{mAP (\%) $\uparrow$}} \\
Loss     & $\theta_{\mathcal{P}_{_{V}}}$ &  $\theta_{\mathcal{P}_{_{L}}}$   & 0.3 & 0.4 & 0.5 & 0.6 & 0.7 & \textbf{Avg.}  \\ 
\midrule
BCE                               & \checkmark &  & 16.7 & 10.8 & 6.4 & 3.5 & 1.8& 7.9 \\
$\mathcal{L}_s$                   &  & \checkmark & 18.1 & 11.7 & 6.9 & 3.7 & 1.7 & 8.4 \\
$\mathcal{L}_z$                   &  & \checkmark & 20.9 & 14.3 & 8.7 & 4.9 & 2.5 & 10.3 \\
\rowcolor{ModelGreen}
$\mathcal{L}_z$ + $\mathcal{L}_s$ & \checkmark  & \checkmark & 20.7 & 14.3 & 8.9 & 5.3 & 2.8 & 10.4 \\
\bottomrule
\end{tabular}%
}
\caption{\textbf{Ablation on learning objective.} Green is \inlineColorbox{ModelGreen}{our configuration}. Results are collected on THUMOS14 (50\%-50\%).}
\label{ablation_loss}
\end{table}

\begin{table}[t]
\small
\centering
\resizebox{.9\columnwidth}{!}{
\begin{tabular}{ll|ccccc|c}
\toprule
\textbf{Method} & \textbf{Split} & \multicolumn{6}{c}{\textbf{mAP (\%) $\uparrow$}} \\
                &                & 0.3 & 0.4 & 0.5 & 0.6 & 0.7 & \textbf{Avg.}  \\ 
\midrule
\methodname$_{\text{no\_sup}}$     & 50:50 & 20.1 & 13.9 & 8.7 & 5.2 & 2.6	& 10.1 \\
\rowcolor{ModelGreen}
\methodname                        & 50:50 & 20.7 & 14.3 & 8.9 & 5.3 & 2.8 & 10.4 \\
\midrule
\methodname$_{\text{no\_sup}}$     & 75:25 & 18.6 & 12.2 & 7.0 & 4.2 & 2.1 & 8.8 \\
\rowcolor{ModelGreen}
\methodname                        & 75:25 & 19.2 & 12.7 & 7.4 & 4.4 & 2.2 & 9.2 \\
\bottomrule
\end{tabular}
}
\caption{\textbf{Ablation on text-guided region suppression.} Green is \inlineColorbox{ModelGreen}{our configuration}. Results are collected on THUMOS14 (50\%-50\%) and on THUMOS14 (75\%-25\%).
\vspace{-5pt}}
\label{ablation_ref}
\end{table}

\noindent{\textbf{Text-guided suppression.}} We assess the efficacy of the final text-guided suppression (Sec.~\ref{textguided}) in Tab.~\ref{ablation_ref}, and report results for the 75:25 and 50:50 splits. The table include results for when we apply this suppression (see Row 2-4) and when we do not (see Row 1-3). Our findings indicate a positive contribution from the suppression in all the settings.

\begin{figure}[!t]
    \centering
    \includegraphics[width=.85\linewidth]{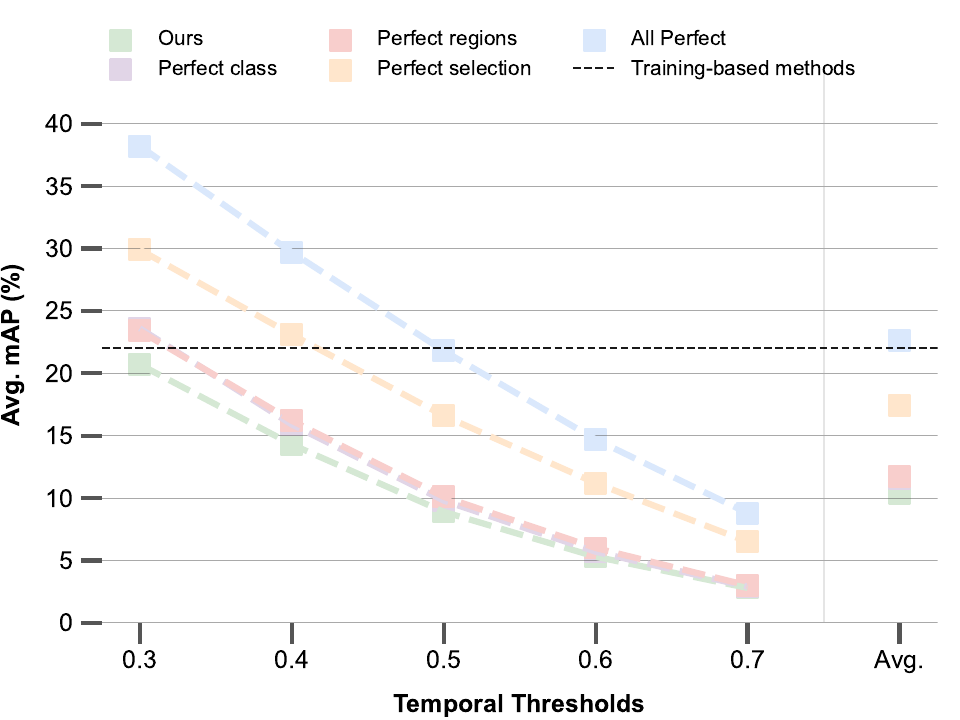}
    \caption{\textbf{Oracle study.} We re-evaluate \inlineColorbox{ModelGreen}{our configuration} with partial perfect information as \inlineColorbox{FunctionPurple}{perfect class} prediction for the pseudo-label, \inlineColorbox{ParamRed}{perfect regions} count selection in the video, and \inlineColorbox{LanguageOrange}{perfect selection} of positive and negative refinement samples. With \inlineColorbox{VisionBlue}{all perfect} mechanisms, we surpass training-based models. 
    \vspace{-15pt}
    }
    \label{fig:ablation_oracle50}

\end{figure}

\noindent{\textbf{Oracle analysis}} 
To validate the potential of our proposed methodology, we conduct an extensive study on top of \methodname. 
These analyses investigate the potential of our TTA method under the relaxation of certain unsupervised constraints. 
Starting from our method design, we identify three components we can replace with oracle information: perfect video-level pseudo-label, perfect region count, and perfect positive selection. We report the performance fluctuation derived from oracle knowledge in Fig.~\ref{fig:ablation_oracle50}.

In the first experiment, we account for an imaginary classifier able to recognize with 100\% accuracy the action from the average representation of the video frames. We replace the proposed pseudo-label $y^\ast$ with such a perfect prediction. As shown in Fig.~\ref{fig:ablation_oracle50}, a better classifier achieves up to +1.2\% mAP. In the second setting, we re-evaluate the performance of our suppression strategy when we select the exact number of action regions $m$ in the video. In this case, after adaptation, we rank the predicted region proposals based on their similarity with the pseudo-label and retain only the first $m$. Similar to the previous analysis, perfect region count results in a relative gain of +1.3\% mAP. Last, we consider the scenario where we retrieve positive samples from intervals encapsulating the video-level pseudo-label and retrieve negative samples from outside such intervals. These results capitalize on a considerable gain in performance, surpassing the improvements on the previous two, with a final score of 17.4\%, \ie, +7\% relative gain. The final experiment consider all the aforementioned constraints relaxation at once. In this configuration, the model performance increase further and achieves 22.6\% on average. Remarkably, these numbers are on on par with state-of-the-art models models evaluated in-domain, without requiring training data or human annotations.
To maintain comparability, the maximum number of adaptation steps used in the main method is kept consistent across all oracle experiments, even if there could potentially be further improvements to the oracle.

\section{Discussion}\label{sec:discussion}
Our research highlighted a shortcoming in existing ZS-TAL literature, indicating an inability of such models for out-of-distribution generalization.
Motivated by this observation, we propose \methodname, a novel approach based on test-time adaptation to fine-tune the model without any training data.
Our method expands a generic VLM, \ie, pre-trained on image data without fine-tuning for TAL, to jointly adapt to video data and learn to localize actions in a zero-shot manner. We achieve all this by only adapting on unlabelled video samples individually. Our experimental evaluation confirms test-time adaptation as a promising direction to 1) calibrate VLMs to solve action localization in videos and 2) mitigate the out-of-distribution generalization problem of current ZS-TAL approaches. Moreover, our study on partial perfect information reveals that test-time adaptation can achieve and surpass current state-of-the-art implementations without training on labeled samples.

\noindent\textbf{Limitations.}
\methodname~relies heavily on good positive and negative samples, which are essential for adequate adaptation to unlabelled data. Our selection protocol tags the frames semantically closer to the video pseudo-label as positives and the ones that are less similar as negatives. Negatives selected in this way, however, may contain completely unrelated concepts such as titles or black screens. Such samples are suboptimal compared to more informative hard negatives. These hard negatives, such as frames that are highly correlated with the actions in videos that are not part of the ground truth regions, provide superior information for the adaptation process. 
Additionally, the video-level pseudo-label restricts to one the number of actions per video, idealizing the real-world setup where multiple actions may appear concurrently.

\noindent\textbf{Potential directions.}
While we propose test-time adaptation to address the out-of-distribution problem of current ZS-TAL models, we acknowledge other directions as viable alternatives, such as cross-domain evaluation protocols (currently studied for video action recognition) or source-free approaches~\cite{xu2022source,zara2023unreasonable}. 
We also believe that VLMs pre-trained for video data will provide a better starting point for temporal visual tasks, similar to a pre-trained action localizer, to adapt at test-time. 
Last, our results also partially highlight that annotated training data is not fundamental to outperform current state-of-the-art ZS-TAL methods. While not explored in this work, we hypothesize that such behavior might be associated with the inherent noise in the label space of action datasets. The existing annotation lacks a well-defined taxonomy, \ie, current approaches must account for a mixture of action verbs, nouns describing actions, and activities (\ie, succession of atomic actions). We advocate for a systematic action taxonomy as a vital future step towards better action-related vision tasks.

\footnotesize\noindent\textbf{Acknowledgment.}~B.L is supported by Leonardo Labs. We acknowledge the CINECA award under the ISCRA initiative, for the availability of HPC resources. This work was also sponsored by EU ISFP PRECRISIS (ISFP-2022-TFI-AG-PROTECT-02-101100539), PNRR ICSC National Research Centre for HPC, Big Data and Quantum Computing (CN00000013) and the FAIR - Future AI Research (PE00000013), funded by NextGeneration EU.

% \listoftodos[List of suggested changes]
{
    \small
    \bibliographystyle{ieeenat_fullname}
    \bibliography{main}
}

\maketitlesupplementary

\noindent In this Supplementary Material, we provide additional quantitative and qualitative results of the proposed \methodname.
In Sec.~\ref{suppprel} we provide details on the preliminary experiment reported in the main manuscript, in Sec.~\ref{suppexp} we discuss per-class results of \methodname, and in Sec.~\ref{suppqual} we show captions generated by the model. The supplementary material is also accompanied by qualitative results in video format that are easily accessible at {\tt\small\url{https://github.com/benedettaliberatori/T3AL}}. These videos can better aid in understanding the results presented in the paper.

\section{Cross-dataset generalization analysis}\label{suppprel}
 
In the experiment reported in Sec. 3 of the main manuscript we consider two state-of-the-art Zero-Shot Temporal Action Localization (ZS-TAL) methods~\cite{Ju2021PromptingVM, stale} that, to the best of our knowledge, are the only works with publicly available code.

For STALE~\cite{stale} we use the model pre-trained on the ActivityNet-v1.3 dataset for the ZS-TAL task. For EffPrompt~\cite{Ju2021PromptingVM}, which does not provide models pre-trained on ZS-TAL datasets, we use a model pre-trained on HMDB51~\cite{6126543} for the video action recognition task. EffPrompt is a two-stage method, \ie, it first detects region proposals and then classifies the obtained regions. For this reason, we employ the same action localizer~\cite{Lin_2021_CVPR} utilized in its first stage to generate action proposals from untrimmed videos, and then use the model pre-trained on trimmed videos to classify the obtained regions. The proposal detector is trained on the original training set of THUMOS14. We use the model pre-trained on THUMOS14 as it is the only one available in the official repository. Consequently, we evaluate its performance for each split using videos from the original test set. The results obtained for this out-of-distribution experiment are reported in Tab.~\ref{tab:crossdataset}, alongside the in-distribution numbers, \ie, models trained and tested on THUMOS14.

STALE trained on ActivityNet-v1.3 is suboptimal when tested on out-of-distribution data. We attribute this performance reduction to different datasets characteristics, as the model trained on ActivityNet-v1.3 learns to predict fewer and longer proposals, but for THUMOS14 regions are generally sparser and shorter. Also EffPrompt shows significantly lower results when evaluated on THUMOS14, despite the pre-training of the proposal detector on the out-of-distribution dataset. This experiment shows that the model is unable to generalize from HMDB51 to THUMOS14 classes.

\begin{table}[h!]
\centering
\resizebox{\columnwidth}{!}{
\begin{tabular}{lcc|ccccc|c}
\toprule
\textbf{Model}  & \textbf{Training Data} & \textbf{Split} & \multicolumn{6}{c}{\textbf{mAP (\%) $\uparrow$}} \\
               & &  & 0.3 & 0.4 & 0.5 & 0.6 & 0.7 & \textbf{Avg.}  \\ 
\midrule
EffPrompt & \colorbox{ModelGreen}{THUMOS14}    & 75:25    &    39.7 & 31.6 & 23.0 & 14.9 & 7.5 & 23.3 \\
EffPrompt & \colorbox{ModelGreen}{THUMOS14}    & 50:50    &    37.2 & 29.6 &  21.6 & 14.0 & 7.2 &  21.9 \\
\midrule
STALE & \colorbox{ModelGreen}{THUMOS14}        & 75:25    &   40.5 & 32.3 &  23.5 & 15.3 & 7.6 & 23.8 \\
STALE & \colorbox{ModelGreen}{THUMOS14}        & 50:50    &   38.3 & 30.7 & 21.2 & 13.8 & 7.0 & 22.2 \\
\midrule
EffPrompt & \colorbox{FunctionPurple}{HMDB51}    & 75:25    & 7.1                    & 5.9                    & 4.5                    & 3.4                    & 2.2                    & 4.6                    \\
EffPrompt & \colorbox{FunctionPurple}{HMDB51}    & 50:50    & 5.4                    & 4.4                    & 3.5                    & 2.7                    & 1.9                    & 3.6                    \\
\midrule
STALE & \colorbox{FunctionPurple}{ActivityNet-v1.3}        & 75:25    & 0.5             & 0.3                    & 0.2                    & 0.2                    & 0.2                    & 0.3                    \\
STALE & \colorbox{FunctionPurple}{ActivityNet-v1.3}        & 50:50    & 1.3             & 0.7                    & 0.6                    & 0.6                    & 0.4                    & 0.7                    \\
\bottomrule
\end{tabular}
}
\caption{\textbf{Cross-dataset generalization.} We show the average mAP, computed at IoU thresholds of [$0.3$:$0.1$:$0.7$], for EffPrompt and STALE trained and tested on \inlineColorbox{ModelGreen}{THUMOS14}, and trained on a \colorbox{FunctionPurple}{different dataset} and tested on THUMOS14. We report results for the 75:25 (75\% seen classes) and 50:50 (50\% seen classes) evaluation settings.}

\label{tab:crossdataset}
\end{table}
\vspace{-5mm}
\section{Experiments}\label{suppexp}

\begin{table}[h]
    \centering
    \resizebox{\columnwidth}{!}{
    \begin{tabular}{lccccc|c}
        \toprule
        \textbf{Class Name} & \multicolumn{6}{c}{\textbf{mAP (\%) $\uparrow$}} \\
                & 0.3 & 0.4 & 0.5 & 0.6 & 0.7 & \textbf{Avg.}  \\ 
        \midrule
        BaseballPitch & 13.2 & 9.4 & 4.4 & 2.5 & 1.6 & 6.2 \\
        BasketballDunk & 23.5 & 13.7 & 7.7 & 4.1 & 1.4 & 10.1 \\
        Billiards & 6.9 & 4.0 & 2.2 & 1.2 & 0.2 & 2.9 \\
        CleanAndJerk & 43.4 & 31.8 & 22.8 & 14.1 & 6.8 & 23.8 \\
        CliffDiving & 33.4 & 23.4 & 14.5 & 9.1 & 4.7 & 17.0 \\
        CricketBowling & 7.6 & 2.6 & 0.9 & 0.3 & 0.1 & 2.3 \\
        CricketShot & 7.1 & 3.3 & 1.2 & 0.5 & 0.2 & 2.5 \\
        Diving & 23.9 & 17.8 & 11.7 & 6.2 & 2.8 & 12.5 \\
        FrisbeeCatch & 8.2 & 4.0 & 1.7 & 0.7 & 0.4 & 3.0 \\
        GolfSwing & 18.1 & 10.6 & 3.6 & 1.3 & 0.8 & 6.9 \\
        HammerThrow & 34.9 & 30.8 & 23.3 & 15.4 & 10.7 & 23.0 \\
        HighJump & 30.3 & 20.3 & 12.2 & 5.9 & 2.9 & 14.3 \\
        JavelinThrow & 29.2 & 21.0 & 13.8 & 8.3 & 4.6 & 15.4 \\
        LongJump & 51.2 & 42.6 & 32.5 & 21.6 & 11.4 & 31.9 \\
        PoleVault & 42.0 & 33.3 & 23.9 & 16.6 & 7.9 & 24.7 \\
        Shotput & 17.1 & 12.2 & 8.0 & 4.8 & 2.8 & 9.0 \\
        SoccerPenalty & 26.7 & 14.0 & 6.7 & 3.0 & 0.8 & 10.3 \\
        TennisSwing & 3.8 & 2.0 & 1.0 & 0.5 & 0.1 & 1.5 \\
        ThrowDiscus & 4.8 & 3.5 & 2.1 & 1.7 & 0.7 & 2.6 \\
        VolleyballSpiking & 19.6 & 14.3 & 7.7 & 4.1 & 2.0 & 9.5 \\
        \bottomrule
    \end{tabular}
    }
    \caption{\textbf{Per-class results on THUMOS14 (50\%-50\%).} Numbers are computed at IoU thresholds of [$0.3$:$0.1$:$0.7$] and averaged across all class splits.}
    \label{supp_50}
\end{table}

\begin{table}[ht]
    \centering
    \resizebox{\columnwidth}{!}{
    \begin{tabular}{lccccc|c}
        \toprule
        \textbf{Class Name} & \multicolumn{6}{c}{\textbf{mAP (\%) $\uparrow$}} \\
                & 0.3 & 0.4 & 0.5 & 0.6 & 0.7 & \textbf{Avg.}  \\ 
        \midrule
BaseballPitch & 12.2 & 7.6 & 2.5 & 1.8 & 1.5 & 5.1 \\
Billiards & 2.0 & 1.4 & 0.3 & 0.2 & 0.0 & 0.8 \\
CleanAndJerk & 29.0 & 20.0 & 11.4 & 5.8 & 3.6 & 14.0 \\
CliffDiving & 37.3 & 25.8 & 16.5 & 10.6 & 5.1 & 19.0 \\
CricketBowling & 7.6 & 2.5 & 1.1 & 0.4 & 0.1 & 2.3 \\
CricketShot & 6.0 & 2.8 & 0.9 & 0.4 & 0.1 & 2.0 \\
Diving & 23.8 & 18.0 & 11.6 & 7.0 & 3.4 & 12.8 \\
FrisbeeCatch & 5.2 & 2.6 & 1.2 & 0.2 & 0.1 & 1.9 \\
GolfSwing & 15.8 & 9.8 & 2.5 & 1.3 & 0.9 & 6.1 \\
HammerThrow & 41.2 & 34.6 & 25.2 & 16.4 & 11.0 & 25.7 \\
HighJump & 32.7 & 20.5 & 12.5 & 5.7 & 2.2 & 14.7 \\
JavelinThrow & 25.0 & 17.2 & 11.3 & 7.6 & 3.4 & 12.9 \\
PoleVault & 50.2 & 37.4 & 25.5 & 18.1 & 8.2 & 27.9 \\
Shotput & 17.0 & 9.0 & 5.0 & 2.6 & 1.8 & 7.1 \\
SoccerPenalty & 26.9 & 15.4 & 7.1 & 2.3 & 1.0 & 10.5 \\
TennisSwing & 3.4 & 2.0 & 1.0 & 0.4 & 0.1 & 1.4 \\
ThrowDiscus & 3.3 & 1.8 & 1.2 & 0.7 & 0.2 & 1.5 \\
VolleyballSpiking & 19.2 & 11.5 & 5.7 & 2.7 & 1.4 & 8.1 \\
\bottomrule
\end{tabular}
}
\caption{\textbf{Per-class results on THUMOS14 (75\%-25\%).} Numbers are computed at IoU thresholds of [$0.3$:$0.1$:$0.7$] and averaged across all class splits.}
\label{supp_75}
\end{table}
\noindent We report per-class results of \methodname~on THUMOS14 for both the evaluation settings, \ie, 50\%-50\% split in Tab.~\ref{supp_50} and 75\%-25\% split in Tab.~\ref{supp_75}. Note that the latter contains only 18 of the total 20 classes as the labels \textit{Basketball dunk} and \textit{Long jump} are not contained in any of the test splits for the 75\%-25\% setting. Following~\cite{Ju2021PromptingVM}, the results are the averages of the individual results obtained across all class splits. Both tables show high variance in performance among the classes. In particular, classes that have less in common with the surrounding scene (\eg, \textit{Clean and jerk}, \textit{Pole vault}, and \textit{Long jump}) exhibit considerably higher results (\eg, 23.8\%, 24.7\%, and 31.9\% avg. mAP on 50:50) compared to classes that share more visual cues with the surrounding context, as observed for \textit{Tennis swing} or \textit{Billiards} (\ie, 1.5\%, 2.9\% avg. mAP on 50:50). We attribute the fact that the model underperforms on videos of class \textit{Tennis swing} to the atomicity of the action: the swing movement bears a subtle difference from a person with a tennis racket in hand who is not actively swinging but is poised and waiting for the ball.  Billiards, instead, serves as an example of an action class that is not atomic but rather encompasses a broad range of potential movements, \eg, holding the billiard cue, striking the ball, or preparing the billiard table. The classes of the datasets contain a mixture of action verbs, nouns describing actions, and activities. The lack of a well-defined taxonomy poses a challenge for TAL methods, as explained in the main manuscript in Sec.~6. 

\section{Qualitative Results}\label{suppqual}
\begin{figure*}[ht]
\centering
    \begin{subfigure}[b]{0.23\textwidth}
        \includegraphics[width=\textwidth]{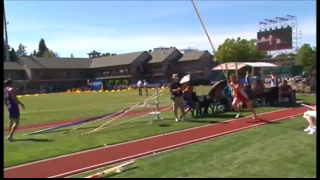}
        \caption*{\centering\texttt{"a \textbf{pole vaulter} is about to take off from the track."}}
        \label{fig:subfig1}
    \end{subfigure}\hspace{0.01\textwidth} 
    \begin{subfigure}[b]{0.23\textwidth}
        \includegraphics[width=\textwidth]{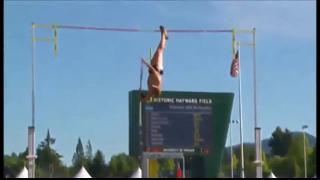}
        \caption*{\centering\texttt{"a \textbf{pole vaulter} is in the air during a competition."}}
        \label{fig:subfig2}
    \end{subfigure}\hspace{0.01\textwidth} 
    \begin{subfigure}[b]{0.23\textwidth}
        \includegraphics[width=\textwidth]{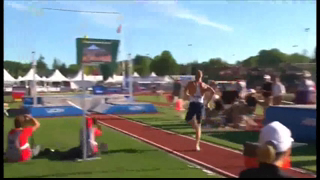}
        \caption*{\centering\texttt{"a man running on a track in a stadium."\newline}}
        \label{fig:subfig3}
    \end{subfigure}\hspace{0.01\textwidth}
    \begin{subfigure}[b]{0.23\textwidth}
    \includegraphics[width=\textwidth]{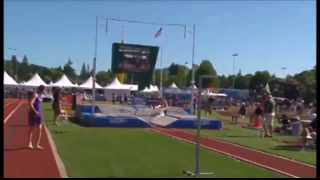}
    \caption*{\centering\texttt{"a \textbf{pole vaulting} event in progress on a field."}}
    \label{fig:subfig4}
    \end{subfigure}
    \caption{\centering Captions generated from frames in video named \texttt{video\_test\_0000793.txt\newline}.}
    \label{fig:cap1}
\centering
    \begin{subfigure}[b]{0.23\textwidth}
        \includegraphics[width=\textwidth]{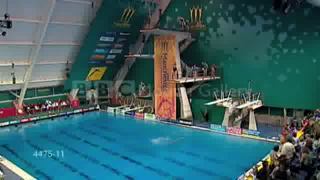}
        \caption*{\centering\texttt{"a swimming pool that has a lot of people in it."}}
        \label{fig:subfig11}
    \end{subfigure}\hspace{0.01\textwidth} 
    \begin{subfigure}[b]{0.23\textwidth}
        \includegraphics[width=\textwidth]{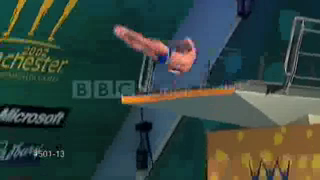}
        \caption*{\centering\texttt{"a man \textbf{diving} off of a \textbf{diving} board."\newline}}
        \label{fig:subfig22}
    \end{subfigure}\hspace{0.01\textwidth} 
    \begin{subfigure}[b]{0.23\textwidth}
        \includegraphics[width=\textwidth]{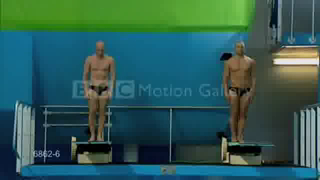}
        \caption*{\centering\texttt{"two men standing on a \textbf{diving} board in the water."}}
        \label{fig:subfig33}
    \end{subfigure}\hspace{0.01\textwidth}
    \begin{subfigure}[b]{0.23\textwidth}
    \includegraphics[width=\textwidth]{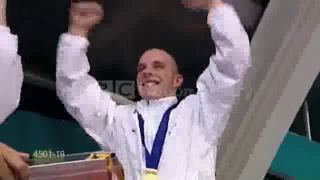}
    \caption*{\centering\texttt{"a man in a white robe is raising his hands."\newline}}
    \label{fig:subfig44}
    \end{subfigure}
    \caption{Captions generated from frames in video named \texttt{video\_test\_0000602.txt}.\newline}
    \label{fig:cap2}
\centering
    \begin{subfigure}[b]{0.23\textwidth}
        \includegraphics[width=\textwidth]{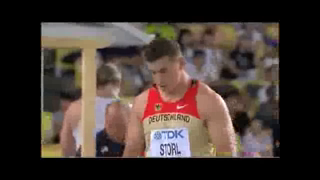}
        \caption*{\centering\texttt{"a man in a red and white shirt is standing in a stadium."}}
        \label{fig:subfig111}
    \end{subfigure}\hspace{0.01\textwidth} 
    \begin{subfigure}[b]{0.23\textwidth}
        \includegraphics[width=\textwidth]{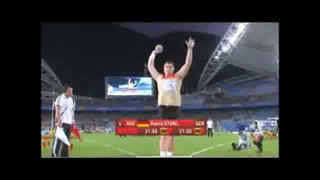}
        \caption*{\centering\texttt{"a man is throwing a \textbf{shot put} in a stadium."\newline}}
        \label{fig:subfig222}
    \end{subfigure}\hspace{0.01\textwidth}
    \begin{subfigure}[b]{0.23\textwidth}
        \includegraphics[width=\textwidth]{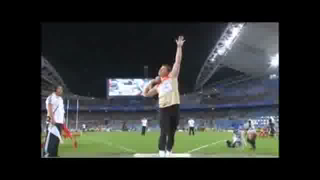}
        \caption*{\centering\texttt{"a man is throwing a \textbf{shot put} in a stadium."\newline}}
        \label{fig:subfig333}
    \end{subfigure}\hspace{0.01\textwidth}
    \begin{subfigure}[b]{0.23\textwidth}
    \includegraphics[width=\textwidth]{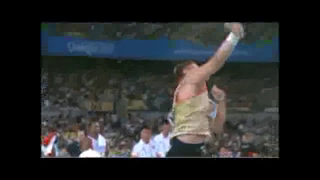}
    \caption*{\centering\texttt{"a man is jumping in the air while holding a frisbee."}}
    \label{fig:subfig444}
    \end{subfigure}
    \caption{Captions generated from frames in video named \texttt{video\_validation\_0000783.txt}.\newline}
    \label{fig:cap3}

\centering
    \begin{subfigure}[b]{0.23\textwidth}
        \includegraphics[width=\textwidth]{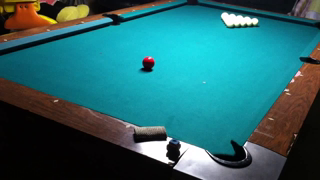}
        \caption*{\centering\texttt{"a pool table with a red ball on it."}\newline}
        \label{fig:subfig1111}
    \end{subfigure}\hspace{0.01\textwidth} 
    \begin{subfigure}[b]{0.23\textwidth}
        \includegraphics[width=\textwidth]{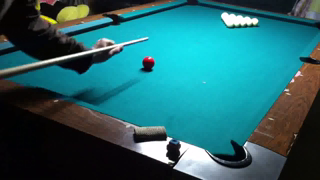}
        \caption*{\centering\texttt{"a person playing pool on a pool table."}\newline}
        \label{fig:subfig2222}
    \end{subfigure}\hspace{0.01\textwidth} 
    \begin{subfigure}[b]{0.23\textwidth}
        \includegraphics[width=\textwidth]{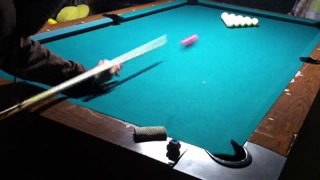}
        \caption*{\centering\texttt{"a pool table with a man playing a game of \textbf{billiards}."}}
        \label{fig:subfig3333}
    \end{subfigure}\hspace{0.01\textwidth}
    \begin{subfigure}[b]{0.23\textwidth}
    \includegraphics[width=\textwidth]{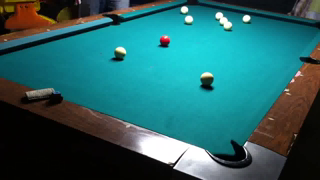}
    \caption*{\centering\texttt{"a green pool table with white and red balls."}}
    \label{fig:subfig4444}
    \end{subfigure}
    \caption{Captions generated from frames in video named \texttt{video\_validation\_0000057.txt}.\newline}
    \label{fig:cap4}

    \centering
    \begin{subfigure}[b]{0.23\textwidth}
        \includegraphics[width=\textwidth]{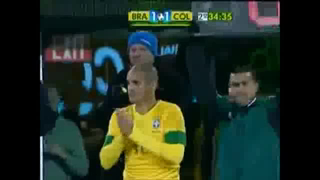}
        \caption*{\centering\texttt{"a \textbf{soccer} player claps his hands in front of a crowd."}}
        \label{fig:subfig11111}
    \end{subfigure}\hspace{0.01\textwidth} 
    \begin{subfigure}[b]{0.23\textwidth}
        \includegraphics[width=\textwidth]{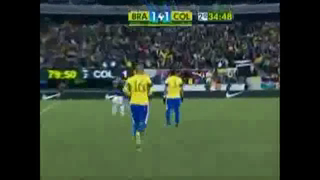}
        \caption*{\centering\texttt{"a group of men playing a game of \textbf{soccer}."}\newline}
        \label{fig:subfig22222}
    \end{subfigure}\hspace{0.01\textwidth} 
    \begin{subfigure}[b]{0.23\textwidth}
        \includegraphics[width=\textwidth]{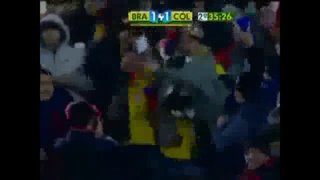}
        \caption*{\centering\texttt{"a crowd of people are gathered together in a stadium."}}
        \label{fig:subfig33333}
    \end{subfigure}\hspace{0.01\textwidth}
    \begin{subfigure}[b]{0.23\textwidth}
    \includegraphics[width=\textwidth]{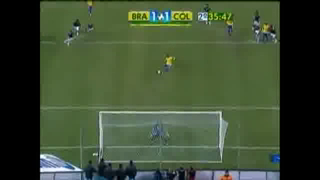}
    \caption*{\centering\texttt{"a \textbf{soccer} player is kicking a ball in front of a crowd."}}
    \label{fig:subfig44444}
    \end{subfigure}
    \caption{Captions generated from frames in video named \texttt{video\_test\_0001153.txt}.}
    \label{fig:cap5}
\end{figure*}

In this section, we show some of the captions generated with CoCa~\cite{yu2022coca} on THUMOS14. It can be seen that captions generated from frames within ground truth regions often contain the ground truth class. Moreover, there are instances where captions contain words related to the annotated class, even when the action is not depicted in the frame, \eg, Fig.~\ref{fig:cap2} containing the word \textit{``diving''} when the individuals in the scene are stationary on the diving board and not engaged in the actual action of diving, or \textit{``pole vaulting''} in Fig.~\ref{fig:cap1} related to a static scene without the performed action. Certain captions may contain words associated with classes different from the ground truth, as illustrated by the example in Fig.~\ref{fig:cap3} where the word \textit{``frisbee''} is present. In this case, the caption shares more semantics with \textit{Frisbee catch} than with the ground truth \textit{Shot put}. There are also cases where words related to the captions (\eg, \textit{``pool''} for the action \textit{Billiards}) are present in captions of images that may or may not depict the action happening, as shown in Fig.~\ref{fig:cap4}. In the case of \textit{Soccer penalty}, the word \textit{``penalty''} is not present in any caption, but the term \textit{``soccer''} is consistently contained in most of them, as shown in Fig.~\ref{fig:cap5}.

\end{document}